\pdfoutput=1

\documentclass[11pt]{article}

\usepackage[]{acl}

\usepackage{times}
\usepackage{latexsym}

\usepackage[T1]{fontenc}

\usepackage[utf8]{inputenc}

\usepackage{microtype}

\usepackage[T1]{fontenc}
\usepackage{type1cm}
\usepackage[american]{babel}
\usepackage{url}
\usepackage{amssymb} 
\usepackage{amsmath}
\usepackage{multirow} 
\usepackage{xspace}
\DeclareMathAlphabet{\mathcalbf}{OMS}{pzc}{b}{n}
\usepackage{microtype}

\newcommand\blfootnote[1]{%
	\begingroup
	\renewcommand\thefootnote{}\footnote{#1}%
	\addtocounter{footnote}{-1}%
	\endgroup
}

\usepackage[pdftex]{graphicx}
\usepackage{tabularx}
\usepackage{booktabs}
\DeclareGraphicsExtensions{.pdf,.ai,.jpg,.png}
\setkeys{Gin}{pagebox=artbox}

\graphicspath{{./coling22-explainability-corpus-figures/}}

\newcommand{\bsfigure}[3][]{%
	\begin{figure}[t]
		\centering
		\includegraphics[#1]{#2}
		\caption{#3}\label{#2}%
 	 \end{figure}
}

\RequirePackage{type1cm}
\RequirePackage{color}
\RequirePackage{soul}
\setstcolor{blue}
\definecolor{violet}{rgb}{0.5,0.0,0.5}
\newsavebox\bscombox
\newcommand{\bscom}[3][]{%
	\sbox{\bscombox}{\fontsize{8}{9}\selectfont#1#2#3}
	\noindent
	\st{#2}{\selectfont
		\color{blue}#3\ifx\\#1\\\else{\fontsize{8}{9}\selectfont\color{violet}[#1]}\fi
	}
}

\definecolor{highlight1}{rgb}{0.95,0.95,0.95}
\definecolor{tgray}{rgb}{0.5,0.5,0.5}
\definecolor{tgray}{rgb}{0.5,0.5,0.5}
\newcommand{\gr}{\color{tgray}}

\begin{document}

\title{``Mama Always Had a Way of Explaining Things So I Could Understand'': A Dialogue Corpus for Learning to Construct Explanations}

\author{Henning Wachsmuth $^*$   \\
Department of Computer Science \\
Paderborn University \\
{\tt henningw@upb.de} \\\And
Milad Alshomary $^*$ \\
Department of Computer Science \\
Paderborn University \\
{\tt milad.alshomary@upb.de} \\}

\date{}

\maketitle

\begin{abstract}
As AI is more and more pervasive in everyday life, humans have an increasing demand to understand its behavior and decisions. Most research on explainable AI builds on the premise that there is one ideal explanation to be found. In fact, however, everyday explanations are co-constructed in a dialogue between the person explaining (the explainer) and the specific person being explained to (the explainee). In this paper, we introduce a first corpus of dialogical explanations to enable NLP research on how humans explain as well as on how AI can learn to imitate this process. The corpus consists of 65 transcribed English dialogues from the Wired video series \emph{5 Levels}, explaining 13 topics to five explainees of different proficiency. All 1550 dialogue turns have been manually labeled by five independent professionals for the topic discussed as well as for the dialogue act and the explanation move performed. We analyze linguistic patterns of explainers and explainees, and we explore differences across proficiency levels. BERT-based baseline results indicate that sequence information helps predicting topics, acts, and moves effectively. 
\blfootnote{$^*$ Both authors contributed equally to this paper.} 
\end{abstract}

\section{Introduction}
\label{sec:introduction}

Explaining is one of the most pervasive communicative processes in everyday life, aiming for mutual understanding of the two sides involved. Parents explain to children, doctors to patients, teachers to students, seniors to juniors---or all the other way round. In explaining dialogues, one side takes the role of the \emph{explainer}, the other the role of the \emph{explainee}. Explainers seek to enable explainees to comprehend a given topic to a certain extent or to perform some action related to it \cite{rohlfing:2021}. This usually implies a series of dialogue turns where both sides request and~provide different information about the topic. In line with the  quote from the movie ``Forrest Gump'' in the title, how an explaining dialogue looks like is strongly affected by the specific explainer and explainee as well as by their interaction.

\bsfigure{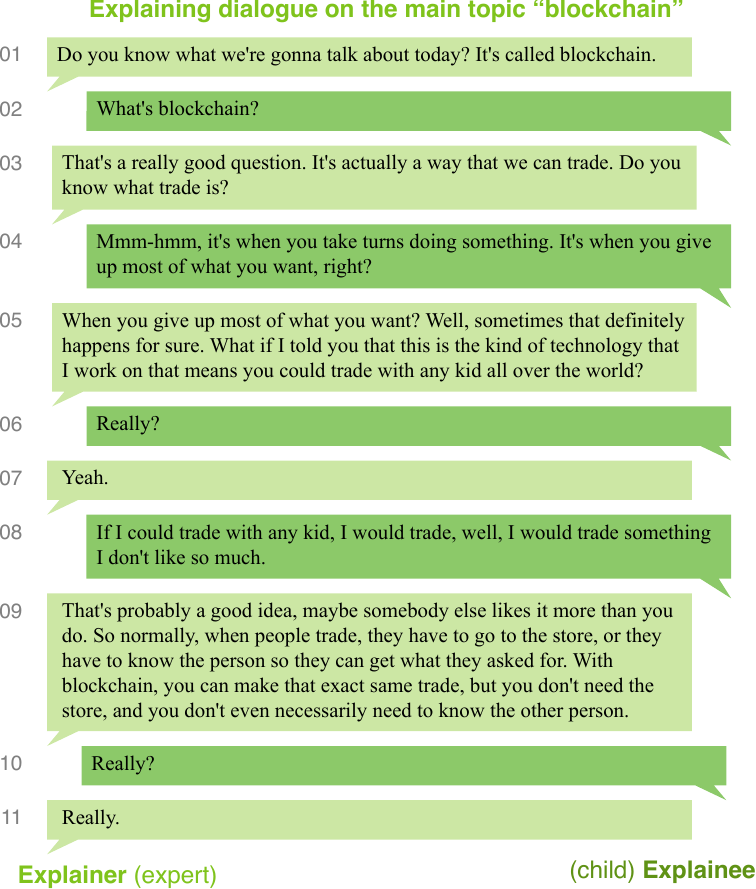}{A short explaining dialogue from the video series \emph{5 Levels}, included in the corpus presented in Section~\ref{sec:data}. Here, an expert explains blockchain to a child.}

Consider the dialogue in Figure~\ref{example.pdf}, where a technology expert explains the basic idea of blockchain to a 5-year old in a controlled setting. Beyond the explanations of the main topic (turns 05 and 09), the dialogue contains an explanation request (02), a test of prior knowledge (03), explanations from the explainee (04), and more. We observe that the explainer's explanations depend on the reaction of the explainee and that their level of depth is most likely adjusted to the explainee's proficiency.

The importance of studying how to explain has become apparent with the rise of research on explainable artificial intelligence, XAI \cite{barriedoarrieta:2020}. As AI finds its way into various aspects of work and private life, humans interacting with respective systems, or being affected by them, have an increasing demand to understand their behavior and decisions. This demand has also been manifested in a \emph{right to explanation} within the EU's General Data Protection Regulation \cite{goodman:2017}. Prior work on XAI largely starts from the premise that an ideal (monological) explanation exists for any behavior or decision, possibly dependent on the explainee at hand \cite{miller:2019}. According to \newcite{rohlfing:2021}, however, real explainability must account for the co-constructive nature of explaining emerging from interaction.

In natural language processing, early work modeled discourse structure of monological explanations \cite{bourse:2012}, and a number of recent approaches generate respective explanations for XAI \cite{situ:2021} and recommendation \cite{li:2021}. In contrast, the language of dialogical explanations is still understudied (details in Section~\ref{sec:relatedwork}). We argue that a better understanding of how humans explain in dialogues is needed, so that XAI can learn to interact with humans.

In this paper, we present a first corpus for computational research on how to explain in dialogues (Section~\ref{sec:data}). The corpus has been created as part of a big interdisplinary research project dealing with the construction of explainability.%
\footnote{Constructing Explainability, https://trr318.upb.de/en}
It consists of 65 transcribed dialogical explanations from the American video series \emph{5 Levels} freely published by the Wired magazine.%
\footnote{5 Levels, https://www.wired.com/video/series/5-levels}
Five dialogues each refer to one of 13 science-related topics (e.g., ``blockchain'' or ``machine learning''). They have the same explainer (an expert on the topic),~but differ in the explainee's proficiency (from child~to~colleague). 

To enable XAI to mimic human explainers, it has to learn what turn to make at any point in a dialogue. 
In discussion with humanities researchers, we model a turn for this purpose by the relation of its \emph{topic} to the main~topic (e.g., subtopic or related topic), its \emph{dialogue act} (e.g., check question or informing statement), and its \emph{explanation move} (e.g., testing prior knowledge or providing an explanation). We segmented the dialogues into a total of 1550 turns, and we let five independent professionals annotate each turn for these three dimensions. 

In Section~\ref{sec:analysis}, we analyze linguistic patterns of explaining dialogues in the annotated corpus. We find clear signals for the explainer's alignment to the explainee's proficiency, such as the avoidance of deviating to related topics towards children. The roles of explainer and explainee are reflected in the varying use of dialogue acts and explanation~moves, possibly stressed by the given setting.

To obtain baselines for the prediction of the three annotated dimensions, we evaluate three variants of BERT \cite{devlin:2019} in 13-topic cross-validation on the corpus (Section~\ref{sec:experiments}). Our results reveal that modeling  sequential dialogue interaction helps predicting~a turn's topic, act, and move effectively. Improvements seem still possible, calling for more sophisticated approaches as well~as~for more explaining dialogue data in the future.%
\footnote{The corpus and the experiment code are freely available here: https://github.com/webis-de/COLING-22}\,\,\,\,

In summary, the contributions of our paper are:
\begin{enumerate}
\setlength{\itemsep}{0pt}
\item
A manually annotated corpus for studying how humans explain in dialogical settings
\item
Empirical insights into how experts explain to explainees of different proficiency levels
\item
Baselines for predicting the topic, dialogue act, and explanation move of dialogue turns \end{enumerate}

\section{Related Work}
\label{sec:relatedwork}

Explainable AI (XAI) largely focuses on the interpretability of learned models from the perspective of scientific completeness \cite{gilpin:2018}. Even though recent works tackle cognitive aspects, such as the trade-off between completeness and compactness \cite{confalonieri:2019}, \newcite{miller:2019} pointed out that this perspective is far away from the understanding of everyday explanations in the social sciences. \newcite{garfinkel:2009} argues that the key is to sort out what the explainer should actually explain, and \newcite{barriedoarrieta:2020} stressed the importance of who is the explainee for XAI. \newcite{rohlfing:2021} built on these works, but reasoned that explanations can only be successful in general, if they are co-constructed in interaction between explainer and explainee. The rationale is that explainees vary in their  motives and needs, and they face different challenges \cite{finke:2022}. The corpus we present serves as a basis for studying the linguistic aspects of the explainer-explainee interaction computationally. 

Natural language language processing (NLP) has notably dealt with the related genre of instructional texts, modeling their structure \cite{fontan:2008}, extracting knowledge \cite{zhang:2012}, comprehending some meaning \cite{yagcioglu:2018}, or generating them \cite{fried:2018}. However, instructional text has a clear procedural style with distinctive surface features \cite{vander:1992}, unlike explanations in general. For tutorial applications, \newcite{jordan:2006} extracted concepts from explanation sentences, whereas \newcite{jansen:2016} studied the knowledge needed for scientific explanations, and \newcite{son:2018} identified causal explanations in social media. Towards a computational understanding of explaining, \newcite{bourse:2012} modeled explanation structure with discourse relations \cite{mann:1988}. In XAI and recommendation contexts, the generation of respective explanations is explored increasingly \cite{situ:2021,li:2021}.

However, our main goal is not to understand how to generate an explanation, but to model how people interact in an explanation process. For annotation, we thus rely on the widely accepted concept of dialogue acts \cite{stolcke:2000,bunt:2010}. Similar has been done for deliberative dialogues by \newcite{alkhatib:2018a}. In addition, we model the \emph{moves} that explainers and explainees make in their interaction, adapting the idea of rhetorical moves, in terms of communicative functions of text segments used to support the communicative objective of a full text \cite{swales:1990}. \newcite{wachsmuth:2017c} proposed task-specific moves for monological arguments, but we are not aware of any work on moves for explanations, nor for dialogical settings.

Hence, we start by compiling data in this paper. Existing related corpora contain tutorial feedback for explanation questions \cite{dzikovska:2012}, answers to non-factoid questions \cite{dulceanu:2018}, and pairs of questions and responses from community question answering platforms \cite{nakov:2017}. Finally, the corpus of \newcite{fan:2019} includes 270k threads from the Reddit forum {\em Explain like I'm Five} where participants explain a concept asked for in simple ways. While all these allow for in-depth analyses of linguistic aspects of explanations, none of them include explaining dialogues with multiple turns on each side. This is the gap we fill with the corpus that we introduce.

\section{Data}
\label{sec:data}

This section introduces the corpus that we created to enable computational research on dialogical explanation processes of humans. We discuss our design choices with respect to the source and annotation, and we present detailed corpus statistics.

\subsection{Explaining Dialogues on Five Levels}

As source data, we decided to rely on explaining dialogues from a controlled setting in which two people explicitly meet to talk about a topic to be explained. While we thereby may miss some interaction behavior found in real-word explanation processes, we expect that such a setting best exhibits explaning dialogue features in their pure form. 

In particular, we acquired the source dialogues in our corpus from \emph{5 Levels}, an American online video series published by the Wired magazine. In each video of the series, one explainer explains a science-related or technology-related topic to five different explainees. The explainer is always an expert on the topic, whereas the explainees increase in terms of (assumed) proficiency on the topic:
\begin{enumerate}
\setlength{\itemsep}{-3pt}
\item
a \emph{child},
\item
a \emph{teenager},
\item
an \emph{undergrad} college sudent, 
\item
a \emph{grad} student, and 
\item
a \emph{colleague} in terms of another expert. 
\end{enumerate}

Every video starts with a few introductory words by the expert, before one dialogue follows the other.%
\footnote{It is noteworthy that the videos seem to have been cut a little, likely for the sake of a concise presentation. We assume that this mainly removed breaks between dialogue turns only. While it limits studying non-verbal interaction in explaining, the effect for textual analyses of the dialogues should be low.}
Transcriptions are already provided in the videos' captions. So far, the first season of the series is available with a total of 17 videos. Table~\ref{table-topics} lists all explained topics (\emph{main topics} henceforth) in these videos, along with explainer information.

\begin{table}[t!]
\small
\renewcommand{\arraystretch}{1}
\setlength{\tabcolsep}{1.3pt}
\centering
\begin{tabular*}{\linewidth}{llll}
\toprule
\bf \#	& \bf Topic				& \bf Explainer			& \bf Expertise \\	
\midrule		
1	& Harmony			& Jacob Collier			& Musician	\\
2	& Blockchain			& Bettina Warburg		& Political scientist \\
3	& Virtual reality			& John Carmack		& Oculus CTO	\\
4	& Connectome			& Bobby Kasthuri		& Neuroscientist \\
5	& Black holes			& Varoujan Gorjian 		& NASA astronomer \\
6	& Lasers				& Donna Strickland		& Professor	 \\
7	& Sleep				& Aric A.\ Prather		& Sleep scientist \\
8	& Dimensions			& Sean Carroll			& Theoret.\ physicist \\
9	& Gravity				& Janna Levin			& Astrophysicist \\
10	& Computer hacking		& Samy Kamkar 		& Security researcher \\
11	& Nanotechnology		& George Tulevski		& Nanotec.\ researcher \\
12	& Origami				& Robert J.\ Lang 		& Physicist \\
13	& Machine learning		& Hilary Mason			& Hidden Door CEO \\
\addlinespace
\gr 14	& \gr CRISPR			& \gr Neville Sanjana		& \gr Biologist \\	
\gr 15	& \gr Memory			& \gr Daphna Shohamy	& \gr Neuroscientist \\
\gr 16	& \gr Zero-knowl.\ proof	& \gr Amit Sahai		& \gr Computer scientist \\
\gr 17	& \gr Black holes		& \gr Janna Levin		& \gr Astrophysicist \\
\bottomrule
\end{tabular*}
\caption{All 17 main topics explained in the \emph{5 Levels} dialogues, along with the explainers and their expertise. The 65 dialogues of the 13 topics listed in black are annotated in our corpus; the rest is provided unannotated.} 
\label{table-topics}
\end{table}


At the time of starting the annotation process discussed below, only 14 of the 17 videos had been accessible, and one of these had partly corrupted subtitles. We thus restricted the annotated corpus to the remaining 13 videos, summing up to 65 dialogues that correspond to a video length of 5.35 hours. Later, we added all dialogues from the other four videos in unannotated form to the corpus. 

Before annotation, we manually segmented each dialogue into its single turns, such that consecutive turns in a dialogue alternate between explainer and explainee. Overall, the 65 dialogues consist of 1550 turns (23.8 turns per dialogue on average), 790 from explainers and 760 from explainees. The turns span 51,344 words (33.1 words per turn). 
On average, an explainer's turn is double as long as an explainee's turn (43.7 vs.\ 22.1 words).
While~the general data size is not huge, we provide evidence in Sections~\ref{sec:analysis} and~\ref{sec:experiments} that it suffices to find common patterns of explanation processes. Limitations emerging from the size are discussed in Section~\ref{sec:conclusion}.%
\footnote{We also extracted the time code (start and end milliseconds) of each segment from the videos, for which one caption is shown. This may serve multimodal studies in the future.}

\subsection{Annotations of Explanatory Interactions}

The corpus is meant to provide a starting point for XAI systems that mimic the explainer's role within dialogical explanation processes. Our annotation scheme supports this purpose and is the result of extensive discussions in our interdisciplinary project with a big team of computer scientists, linguists, psychologists, and cognitive scientists. Where possible, we followed the literature, but the lack of research on human interaction in explaining (see Section~\ref{sec:relatedwork}) made us extend the state of the art in different respects.

In particular, we focus on turn-level category labels that capture the basic behavior of explainers and explainees in explaining dialogues. Our scheme models the three dimensions of dialogue turns that we agreed on to be needed for a computational understanding of the behavior:
\begin{itemize}
\setlength{\itemsep}{-2pt}
\item
the relation of a turn's \emph{topic} to the main topic,
\item
the \emph{dialogue act} performed in the turn, and
\item
the \emph{explanation move} made through the turn.
\end{itemize}

\noindent
We discuss the labels considered for each of the three annotation dimensions in the following. Since all labels apply to both explainer and explainee in principle, we refer to a speaker and a listener below.

\paragraph{Topic}

Even though the dialogues we target have one defined main topic to be explained, what is explained in specific turns may vary due to the dynamics of explaining interaction \cite{garfinkel:2009}. Since we seek to learn how to explain in general rather than any specificities of the concrete 13 main topics in the corpus, we abstract from the latter, modeling only the relation of the topic discussed in a turn to the dialogue's main topic. In particular, a turn's topic may be annotated as follows:
\begin{enumerate}
\setlength{\itemsep}{0pt}
\item[t$_1$]
\emph{Main topic.} The main topic to be explained;
\item[t$_2$]
\emph{Subtopic.} A specific aspect of the main topic;
\item[t$_3$]
\emph{Related topic.} Another topic that is related to the main topic;
\item[t$_4$] 
\emph{No/Other topic.} No topic, or another topic that is unrelated to the main topic.
\end{enumerate}

\paragraph{Dialogue Act}

To model the communicative functions of turns in dialogues, we follow the literature \cite{bunt:2010}, starting from the latest version of the ISO standard taxonomy of dialogue acts.%
\footnote{DIT++ Taxonomy of Dialogue Acts, https://dit.uvt.nl}
In explaining, specific dialogue acts are in the focus, though. In collaboration with the interdisciplinary team, we selected a subset of 10 acts that capture communication on a level of detail that is specific enough to distinguish key differences, but abstract enough to allow finding recurring patterns:
\begin{enumerate}
\setlength{\itemsep}{0pt}
\item[d$_1$]
\emph{Check question.} Asking a check question;
\item[d$_2$]
\emph{What/How question.} Asking a what question or a how question of any kind;
\item[d$_3$]
\emph{Other question.} Asking any other question;
\item[d$_4$] 
\emph{Confirming answer.} Answering a question with confirmation;
\item[d$_5$]
\emph{Disconfirming answer.} Answering a question with disconfirmation;
\item[d$_6$] 
\emph{Other answer.} Giving any other answer;
\item[d$_7$]
\emph{Agreeing statement.} Conveying agreement on the last utterance of the listener;
\item[d$_8$]
\emph{Disagreeing statement.} Conveying disagreement accordingly;
\item[d$_9$]
\emph{Informing statement.} Providing information with respect to the topic stated in the turn;
\item[d$_{10}$] 
\emph{Other.} Performing any other dialogue act.
\end{enumerate}

\paragraph{Explanation Move}

Finally, we aim to understand the explanation-specific moves that explainers and explainees make to work together towards a successful explanation process. Due to the lack of models of explaining dialogues (see Section~\ref{sec:relatedwork}, we started from recent theory of explaining \cite{rohlfing:2021}. Based on a first inspection of a corpus sample, we established a set of 10 explanation moves that a speaker may make~in~the process, at a granularity similar to the dialogue acts:%
\footnote{We decided to leave a distinction of different explaining types (such as causal or analogy-based explanations) to future work, as it does not match the level of detail in our scheme.}
\begin{enumerate}
\setlength{\itemsep}{0pt}
\item[e$_1$]
\emph{Test understanding.} Checking whether the listener understood what was being explained;
\item[e$_2$]
\emph{Test prior knowledge.} Checking the listener's prior knowledge of the turn's topic;
\item[e$_3$]
\emph{Provide explanation.} Explaining any concept or a topic to the listener;
\item[e$_4$] 
\emph{Request explanation.} Requesting any explanation from the listener;
\item[e$_5$]
\emph{Signal understanding.} Informing the listener that their last utterance was understood;
\item[e$_6$]
\emph{Signal non-understanding.} Informing the listener that the utterance was not understood; 
\item[e$_7$]
\emph{Providing feedback.} Responding qualitatively to an utterance by correcting errors or similar;
\item[e$_8$] 
\emph{Providing assessment.} Assessing the listener by rephrasing their utterance or giving a hint;
\item[e$_9$]
\emph{Providing extra info.} Giving additional information to foster a complete understanding;
\item[e$_{10}$] 
\emph{Other.} Making any other explanation move.
\end{enumerate}

We note the hierarchical nature of the scheme with respect to dialogue acts and explanations; for example, d$_1$--d$_3$ could be merged as well as e$_1$--e$_2$. While some acts and moves are much more likely to be made by an explainer or an explainee, we did not restrict this to avoid biasing the annotators.%
\footnote{For dialogue acts d$_3$, d$_6$, and d$_{10}$ as well as explanation move e$_{10}$, the annotators had to name the label in free text. We provide these as part of the corpus, we give individual examples of other moves and acts in Section~\ref{sec:analysis}.}

\subsection{Crowd-based Annotation Process}

The restriction of the annotations to a manageable number of turn-level labels was also made to make the annotation process simple enough to carry it out with independent people. In particular, we hired five freelancers, working as content editors and annotators on the professional crowdworking platform \emph{Upwork}. All were native speakers of English with a 90\%+ job success rate on the platform. We clarified the task individually with each of them. 

We provided guidelines based on the definitions above, along with general explanations and some examples. Using Label Studio,%
\footnote{Label Studio, https://labelstud.io} we developed a task-specific user interface where each dialogue was shown as a sequence of turns and one label of each dimension could be assigned to a turn (if multiple labels seemed appropriate, the best fitting one). Each annotator labeled all 1550 turns.~We~paid \$ 1115 for an overall load of 85~hours, that is, \$~13.12 per hour on average (with minor differences for annotators due to bonuses and varying durations).

\paragraph{Agreement}

In terms of the conservative measure Fleiss' $\kappa$, the inter-annotator agreement among all five was 0.35 for the topic, 0.49 for dialogue acts, and 0.43 for explanation moves. While these values indicate moderate agreement only, they are in line with related subjective labeling tasks of short texts such as news sentences \cite{alkhatib:2016b} and social media arguments \cite{habernal:2018a}. Moreover, we exploited the multiple labels we have per turn to consolidate reliable annotations, as described in the following.


\paragraph{Output Annotations}

For consolidation, we rely on MACE \cite{hovy:2013}, a widely used technique for grading the reliability of crowdworkers based on their agreement with others. The MACE competence scores of the annotators suggest that all did a reasonable job in general, lying in the ranges 0.30--0.76 (topic), 0.58--0.82 (dialogue acts), and 0.45--0.85 (explanation moves) respectively. We applied MACE' functionality to derive one aggregate output label for each dimension from the five annotations weighted by competence scores.

\subsection{The Wired Explaining Dialogue Corpus}

\begin{table}[t!]
\small
\renewcommand{\arraystretch}{1}
\setlength{\tabcolsep}{1pt}
\centering
\begin{tabular*}{\linewidth}{l@{}l@{\hspace*{-0.3cm}}rrc@{}rrc@{}rr}
\toprule
 		&					& \multicolumn{2}{r}{\bf Explainer}	&& \multicolumn{2}{r}{\bf Explainee}	&& \multicolumn{2}{r}{\bf Total}  \\
									\cmidrule(l@{4pt}r@{0pt}){3-4}		\cmidrule(l@{4pt}r@{0pt}){6-7}		\cmidrule(l@{4pt}r@{0pt}){9-10}	
\multicolumn{2}{l}{\bf Label}		& \bf \#	& \bf  \%				&& \bf \#	& \bf  \%				&& \bf \#	& \bf  \% \\	
\midrule		
t$_{1}$	& Main topic			&\quad \bf 301	&\bf 38.1	&&\quad 129	&17.0	&& \quad430	& 27.7	\\
t$_{2}$	& Subtopic			&52	&6.6	&&36	&4.7	&& 88	& 5.7	\\
t$_{3}$	& Related topic			&142	&18.0	&&118	&15.5	&& 260	& 16.8	\\
t$_{4}$	& Other/No topic		&295	&37.3	&& \bf 477	&\bf 62.8	&& \bf 772	& \bf 49.8	\\
\midrule
d$_{1}$	& Check question		&183	&23.2	&&62	&8.2	&& 245	& 15.8	\\
d$_{2}$	& What/How question	&77	&9.7	&&38	&5.0	&& 115	& 7.4	\\
d$_{3}$	& Other question		&3	&0.4	&&10	&1.3	&& 13	& 0.8	\\
d$_{4}$	& Confirming answer		&14	&1.8	&&40	&5.3	&& 54	& 3.5	\\
d$_{5}$	& Disconfirm. answer	&3	&0.4	&&21	&2.8	&& 24	& 1.5	\\
d$_{6}$	& Other answer			&2	&0.3	&&23	&3.0	&& 25	& 1.6	\\
d$_{7}$	& Agreeing statement	&75	&9.5	&&190	&25.0	&& 265	& 17.1	\\
d$_{8}$	& Disagree. statement	&2	&0.3	&&10	&1.3	&& 12	& 0.8	\\
d$_{9}$	& Informing statement	&\bf 391	&\bf 49.5	&&\bf 305	&\bf 40.1	&& \bf 696	& \bf 44.9	\\
d$_{10}$	& Other				&40	&5.1	&&61	&8.0	&& 101	& 6.5	\\
\midrule
e$_{1}$	& Test understanding	&56	&7.1	&&0	&0.0	&& 56	& 3.6	\\
e$_{2}$	& Test prior knowledge	&111	&14.1	&&1	&0.1	&& 112	&7.2	\\
e$_{3}$	& Provide explanation	&\bf 409	&\bf 51.8	&&\bf 270	&\bf 35.5	&& \bf 679	&\bf 43.8	\\
e$_{4}$	& Request explanation 	&47	&5.9	&&95	&12.5	&& 142	&9.2	\\
e$_{5}$	& Signal understanding 	&37	&4.7	&&104	&13.7	&& 141	&9.1	\\
e$_{6}$	& Signal non-underst.	&1	&0.1	&&16	&2.1	&& 17	&1.1	\\
e$_{7}$	& Provide feedback		&61	&7.7	&&224	&29.5	&& 285	&18.4	\\
e$_{8}$	& Provide assessment	&10	&1.3	&&1	&0.1	&& 11	&0.7	\\
e$_{9}$	& Provide extra info		&26	&3.3	&&22	&2.9	&& 48	&3.1	\\
e$_{10}$	& Other				&32	&4.1	&&27	&3.6	&& 59	& 3.8	\\
\midrule
$\Sigma$	&					&790	&100.0	&&760	&100.0	&&1550	&100.0	\\
\bottomrule
\end{tabular*}
\caption{Corpus distribution of annotated topics (t$_1$--t$_4$), dialogue acts (d$_1$--d$_{10}$), and explanation moves (e$_1$--e$_{10}$) separately for explainer and explainee turns and in total. Per type, the highest value in a column is marked bold.} 
\label{table-annotations}
\end{table}

Table~\ref{table-annotations} presents detailed general statistics of the three annotation dimensions. More insights into the distribution of annotations across proficiency levels follow in Section~\ref{sec:analysis}.

With respect to topic (t$_1$--t$_4$), about half of all turns explicitly discuss the \emph{main topic} (27.7\%), a \emph{subtopic} (5.7\%), or a \emph{related topic} (16.8\%). Explainees much more often mention none of these (62.8\% vs.\ 37.3\%), underlining the leading role of the explainer in dialogue setting.

For dialogue acts (d$_1$--d$_{10}$), we see that, quite intuitively, \emph{informing statements} (44.9\%) are dominant in explaining dialogues on both sides (explainer 49.5\%, explainee 40.1\%). However, also \emph{agreeing statements} (17.1\%) as well as \emph{check questions} (15.8\%) play an important role. The low frequency of \emph{other questions} (0.8\%) and \emph{other} (6.5\%) suggests that the selected set of dialogue acts cover well what happens in the given kind of dialogues, even though our annotators identifid sum acts, such as \emph{disagreeing statements} (0.8\%), rarely only.%
\footnote{Notable examples of other dialogue acts the annotators observed include \emph{greetings} (e.g., ``Hi, are you Bella?''), \emph{casual chat} (``What do you do?''), and \emph{gratitude} (``Thank you.'').}

Similar holds for the explanation moves (e$_1$--e$_{10}$): only 3.8\% of all 1550 turns belong to \emph{other}.%
\footnote{Here, other cases include \emph{inquiry} (``Hi, are you Bella'') and \emph{introduction} (``Bella, I'm George, nice to meet you.'').}
As expected, the core of explaining is to \emph{provide explanations} (43.8\%), also explainees do so in 270 turns (35.5\%). Besides, they often \emph{provide feedback} (29.5\%). Explainers rather \emph{test prior knowledge} (14.1\%) and \emph{test understanding} often (7.1\%), but also provide feedback sometimes (7.7\%).

\section{Analysis}
\label{sec:analysis}

One main goal of the presented corpus is to learn how humans explain in dialogical settings. This section analyzes commonalities and differences regarding meta-information available in the corpus.

\subsection{Explaining across Proficiency Levels}

First, we explore to what extent explaining differs depending on the proficiency of the explainee. Figure~\ref{distribution-by-proficiency.pdf} shows the distributions of the three annotated dimensions separately for the five given explainee levels. For dialogue acts and explanation moves, we distinguish only the most frequent labels and merge all others into a class \emph{rest}.

\bsfigure{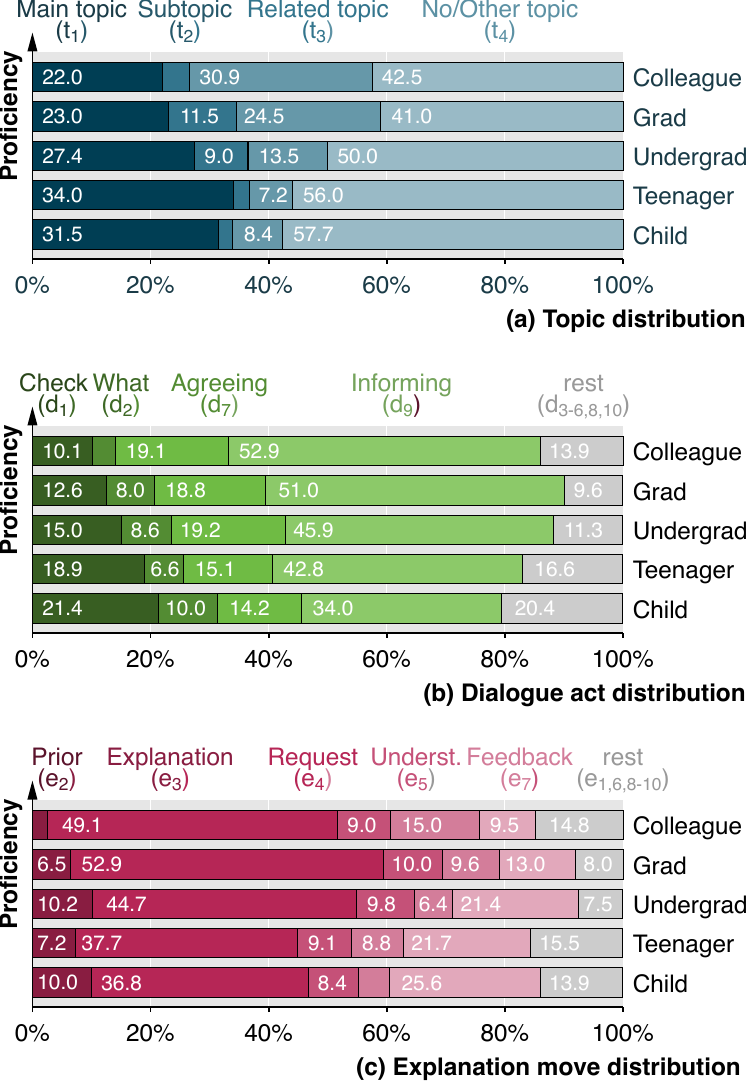}{Distribution of topic, discourse act, and explanation act annotations in the corpus, depending on the proficiency of the explainee (from \emph{Child} to \emph{Colleague}).}

With respect to topic, we see that particularly the discussion of \emph{related topics} grows notably with the explainee's proficiency, from 8.4\% of all annotations for children to 30.9\% for colleagues. Conversely, the \emph{main topic} is mentioned less in dialogues with more proficient explainees; the same holds for \emph{no/other topic}. \emph{Subtopics} are considered mainly with grads (11.5\%) and undergrads (9.0\%), possibly related to the way they learn.

For dialogue acts, the key difference lies between the proportion of \emph{informing statements} and the number of questions asked (d$_1$ and $d_2$). Whereas the former monotonously goes up from 34.0\% (child) to 52.9\% (colleague), particularly the use of \emph{check questions} is correlated inversely with proficiency, used mainly to test prior knowledge and to check understanding. A similar behavior can be observed for explanation moves. There, \emph{providing feedback} shrinks from 25.6\% to 9.5\%, while \emph{providing explanations} mostly grows, with peak at grads (52.9\%). In contrast, how often people \emph{request explanations} remains stable across proficiency levels.

\begin{table}[t!]
\small
\renewcommand{\arraystretch}{1}
\setlength{\tabcolsep}{2pt}
\centering
\begin{tabular*}{\linewidth}{l@{\hspace*{-0.7cm}}rr@{\quad}r}
\toprule
\bf Topic Sequences 							& \bf Explainer 		& \bf Explainee 		& \bf Total  \\
\midrule				
(Main, Rel, Main)          		  				& \bf 24.6\%			& 7.7\%			& \bf 15.4\%	\\
(Main, Rel, Main, Rel, Main)					& --\phantom{\%}	& --\phantom{\%} 	& 7.7\% 	\\	
(Main)                                           					& 12.3\%			& \bf 18.5\%			& 6.2\%	\\
(Rel, Main, Rel, Main, Rel, Main) 					& --\phantom{\%}  	& --\phantom{\%} 	& 4.6\%	\\	
(Main, Rel)                                 					& 3.1\%			& 10.8\%			& 4.6\%	\\	
(Rel, Main, Rel, Main)                  				& 3.1\%			& --\phantom{\%}	& 3.1\%	\\	
(Main, Sub, Main)                               				& --\phantom{\%}  	& --\phantom{\%} 	& 3.1\%	\\	
(Main, Sub, Main, Rel, Main)               			& 4.6\%			& 3.1\%			& 3.1\%	\\	
\bottomrule
\end{tabular*}
\caption{Relative frequencies of all recurring sequences of \emph{main}, \emph{sub}, and \emph{rel}ated topic in the corpus' dialogues and in the explainers and explainees' parts alone.} 
\label{table-flows}
\end{table}

\subsection{Interactions of Topics, Moves, and Acts}

Interactions of the annotated dimensions happen between the turns and within a turn. We analyze one example of each here, and, due the limited data size, we look at topics separately from dialogue act and explanation moves.

Inspired by the flow model of \newcite{wachsmuth:2017c}, Table~\ref{table-flows} shows all eight sequences of topics that occur more than once among the 65 dialogues. Each sequence shows the ordering of topics being discussed, irrespective of how often each topic is mentioned in a row. Most dialogues start and end with the main topic, often in alternation with related topics, such as \emph{(Main, Rel, Main)} in 15.4\% of all cases (sometimes also with subtopics). The ordering of what \emph{explainers} talk about is similar, whereas \emph{explainees} often focus on the main topic only (18.5\%).

Table~\ref{table-pairs} lists the top-10 pairs of acts and moves. \emph{Informing statements} that \emph{provide explanations} are most common across both explainers (45.9\%) and explainees (31.3\%). \emph{Agreeing statements} (d$_{7}$) and \emph{check questions} (d$_{1}$) cooccur with multiple moves, and especially \emph{providing feedback} happens via different dialogue acts. As expected in the given setting, explainees never check for prior knowledge or understanding (d$_{1}$/e$_{2}$, d$_{1}$/e$_{1}$). Instead, they agree by providing feedback or signaling understanding (d$_{7}$/e$_{7}$, d$_{7}$/e$_{5}$) much more often than explainers.

\begin{table}[t!]
\small
\renewcommand{\arraystretch}{1}
\setlength{\tabcolsep}{2pt}
\centering
\begin{tabular*}{\linewidth}{l@{\,}l@{\hspace*{-0.55cm}}rr@{\quad}r}
\toprule
\bf Labels			& \bf Act/Move Pair 			& \bf Explainer 		& \bf Explainee 		& \bf Total  \\
\midrule				
d$_{9}$/e$_{3}$	& Informing/Explanation 		& \bf 45.9\%		& \bf 31.3\%		& \bf 38.8\%	 \\
d$_{7}$/e$_{7}$	& Agreeing/Feedback   		& 3.9\%			& 14.2\%			& 9.0\%	 \\
d$_{7}$/e$_{5}$	& Agreeing/Understanding    	& 3.5\%			& 9.1\%			& 6.3\%	 \\
d$_{1}$/e$_{2}$	& Check/Prior    			& 10.5\%			& --\phantom{\%}	& 5.4\%	 \\
d$_{1}$/e$_{4}$	& Check/Request      		& 2.7\%			& 6.8\%			& 4.7\%	 \\
d$_{2}$/e$_{4}$	& What/Request    			& 3.0\%			& 4.5\%			& 3.7\%	 \\
d$_{10}$/e$_{10}$	& Other/Other     			& 2.8\%			& 2.6\%			& 2.7\%	 \\
d$_{1}$/e$_{1}$	& Check/Understanding      	& 5.1\%			& --\phantom{\%}	& 2.6\%	 \\
d$_{4}$/e$_{7}$	& Confirming/Feedback     	& 1.4\%			& 3.7\%			& 2.5\%	 \\
d$_{9}$/e$_{7}$	& Informing/Feedback   		& 0.5\%			& 4.2\%			& 2.3\%	 \\
\bottomrule
\end{tabular*}
\caption{Relative frequencies of the ten most frequent pairs of dialogue act and explanation move in the corpus and the differences for explainers and explainees.} 
\label{table-pairs}
\end{table}

\subsection{Language of Explainers and Explainees}

Finally, we investigate basic differences in the language of the two sides: We determine the words that are often used by explainers (at least 0.1\% of all words) and rarely by explainees, or vice versa.\,\,\,

Table~\ref{table-words} presents the 10 most specific words on each side. Aside from some topic-specific words (e.g., ``light''), the explainer's list includes typical  words used in meta-language, as in this explanation to a teenager: ``I \emph{want} to know if you agree, sleep is the coolest \emph{thing} you've ever heard of.'' On the explainee's side, we find multiple reactive words, such as ``oh'' and ``interesting'', but also indicators of vagueness, as in this colleague's response to an explanation of hacking: ``So all kind of older logic and \emph{stuff like} that. So, I \emph{mean}, it's sort of based on, \emph{like}, you're presented the little MUX chip.'' 


\begin{table}[t!]
\small
\renewcommand{\arraystretch}{1}
\setlength{\tabcolsep}{3pt}
\centering
\begin{tabular*}{\linewidth}{l@{\,}rrcl@{\,}rr}
\toprule
\multicolumn{3}{c}{\bf Explainer}	&	& \multicolumn{3}{c}{\bf Explainee}  \\
\cmidrule(l@{2pt}r@{2pt}){1-3}		\cmidrule(l@{2pt}r@{2pt}){5-7}
\bf Word	& \bf Frequency	& \bf Ratio	&	& \bf Word	& \bf Frequency	& \bf Ratio \\	
\midrule															
here		& 0.16\%	& 4.20		& 	& yes		& 0.21\%	& 5.12 \\
around	& 0.12\%	& 4.03		& 	& mean		& 0.14\%	& 4.20 \\
space	& 0.24\%	& 3.32		& 	& stuff		& 0.11\%	& 3.11 \\
light		& 0.18\%	& 2.96		& 	& oh			& 0.16\%	& 2.75 \\
earth		& 0.10\%	& 2.65		& 	& yeah		& 0.65\%	& 2.70 \\
us		& 0.15\%	& 2.39		& 	& many		& 0.12\%	& 2.39 \\
want		& 0.14\%	& 2.28		& 	& interesting	& 0.12\%	& 2.11 \\
going	& 0.22\%	& 2.19		& 	& well		& 0.21\%	& 1.94 \\
point		& 0.11\%	& 2.11		& 	& like		& 1.10\%	& 1.85 \\
thing		& 0.18\%	& 1.93		& 	& no			& 0.18\%	& 1.83 \\
\bottomrule
\end{tabular*}
\caption{The top-10 words used specifically by explainers and explainees, respectively, along with the relative frequency (minimum 0.1\%) and specificity ratio (e.g., explainees say ``yes'' 5.12 times as often as explainers).} 
\label{table-words}
\end{table}

\section{Experiments}
\label{sec:experiments}

The second goal of the corpus is to serve the creation of XAI systems that mimic human explainers. As an initial endeavor, this section reports on baseline experiments on the computational prediction of topics, dialogue acts, and explanation moves.

\subsection{Experimental Setup}

We evaluate three models based on BERT \cite{devlin:2019}, along with a simple majority baseline, for predicting each dialogue turn dimension in 13-fold cross-topic validation: For each main topic, we trained one model on the other 12 topics and tested it against the labels of the respective dimension. We average the resulting F$_1$-scores over all 13 folds.%
\footnote{All models start from the \texttt{bert-based-uncased}, and are trained with a learning rate of $2e^{-5}$ and a batch size of 4.}
Figure~\ref{bert-variants.pdf} illustrates the three BERT variants.

\paragraph{BERT-basic} 

The first model simply adds a classification head to BERT. It takes as input the dialogue's main topic and the turn's text, $x_i$ (separated by \texttt{[SEP]}), as well as the label $y_i$ to predict (topic $t_i$, dialogue act $d_i$, or explanation move $e_i$). We trained the model for five epochs, optimizing its F$_1$-score on the turns of two main topics. We balanced the training set using oversampling to prevent the model from only predicting the majority label.

\paragraph{BERT-sequence} 

Turns made in explaining dialogues depend on previous turns, for example, a conclusion on the \emph{main topic} may be preceded by a \emph{related topic} (see Table~\ref{table-flows}). In the second model, we exploit such dependencies with turn-level sequence labeling: Given the sequence $(x_1, \ldots, x_n)$ of all turns in a dialogue, the input to predicting a label $y_i$ of $x_i$ is the turn's history $(x_1, \ldots, x_{i-1})$~along with all previously predicted labels $(y_1, \ldots, y_{i-1})$ of the same dimension. For each turn, we encode the history in a \texttt{CLS} embedding with BERT.
Then, we pass all labels and \texttt{CLS} embeddings through a CRF layer to model the label's dependencies.

\bsfigure{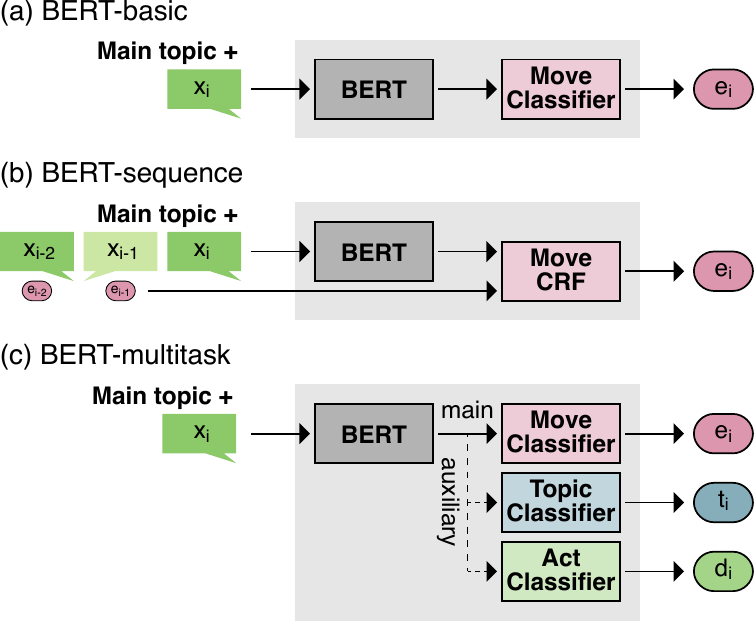}{Sketch of the three evaluated models, here for predicting a turn's explanation move, e$_i$: (a)~\emph{BERT-basic} labels a turn in isolation. (b)~\emph{BERT-sequence} takes the labels of previous turns into account. (c)~\emph{BERT-multitask} classifies all three turn dimensions simultaneously.}

\begin{table}[t]%
	\centering
	\small
	\setlength{\tabcolsep}{1.5pt}%
	\renewcommand{\arraystretch}{1}
	\begin{tabular}{l@{$\!\!\!$}rrrrcr}
		\toprule
					& \bf Main	& \bf Sub- 	& \bf Related	& \bf No/Oth.	 && \bf Macro	\\
		\bf  Approach 	& \bf T.\ ($t_1$) & \bf T.\ ($t_2$) & \bf  T.\ ($t_3$) & \bf T.\ ($t_4$) && \bf F$_1$-Score \\
		\midrule		
		BERT-basic	 	& 0.58 	   & 0.11 	  & \bf 0.44 & \bf 0.89 && 0.51\\
		BERT-sequence	& \bf 0.61 & \bf 0.13 & \bf 0.44 & \bf 0.89 && \bf 0.52\\
		BERT-multitask	& 0.43 	   & 0.04 	  & 0.36 	 & 0.81 	&& 0.41\\
		\addlinespace
		Majority baseline  & 0.00 &  0.00 &  0.00 & 0.66 && 0.17\\
		\bottomrule
	\end{tabular}
	\caption{Topic prediction results: The F$_1$-scores of the evaluated BERT models for each considered relation to the main topic, $t_1$--$t_{4}$, as well as the macro-averaged F$_1$-score. The best value in each column is marked bold.}
	\label{table-results-topic}
\end{table}

\begin{table*}[t]%
	\centering
	\small
	\setlength{\tabcolsep}{3pt}%
	\renewcommand{\arraystretch}{0.985}
	\begin{tabular}{l@{}rrrrrrrrrrcr}
		\toprule
						& \bf Check	& \bf What/H. 	& \bf Other	& \bf Confirm.	& \bf Disconf.	& \bf Other	& \bf Agree.	& \bf Disagr.	& \bf Inform.	& \bf Other && \bf Macro	\\
		\bf  Approach 		& \bf Q.\ ($d_1$) & \bf Q.\ ($d_2$) & \bf  Q.\ ($d_3$) & \bf A.\ ($d_4$) & \bf A.\ ($d_5$) & \bf A.\ ($d_6$) & \bf St.\ ($d_7$) & \bf St.\ ($d_8$) & \bf St.\ ($d_9$) & ($d_{10}$)  && \bf F$_1$-Score \\
		\midrule		
		
		BERT-basic 	    & \bf 0.76 & \bf 0.73 & 0.00 & 0.33 & \bf 0.67 & 0.00 & 0.51 & 0.00 & \bf 0.87 & 0.57 && 0.44 \\
		BERT-sequence	& \bf 0.76 & 0.72 & 0.00 & \bf 0.35 & \bf 0.67 & 0.00 & \bf **0.69 & 0.00 & \bf 0.87 & \bf 0.61 && \bf 0.47\\
		BERT-multitask	& 0.54 & 0.49 & 0.00 & 0.29 & 0.59 & 0.00 & 0.53 & \bf 0.09 & 0.84 & 0.44 && 0.38\\
		\addlinespace
		Majority baseline	& 0.00 & 0.00 &  0.00 &  0.00 & 0.00 & 0.00 &  0.00 & 0.00 &  0.62 &  0.00 && 0.06  \\
		\bottomrule
	\end{tabular}
	\caption{Dialogue act prediction results: The F$_1$-scores of the evaluated BERT models for each considered dialogue act, $d_1$--$d_{10}$, as well as the macro-averaged F$_1$-score. The best value in each column is marked bold.}
	\label{table-results-dialogue-act}
\end{table*}

\begin{table*}[t]%
	\centering
	\small
	\setlength{\tabcolsep}{2.8pt}%
	\renewcommand{\arraystretch}{0.985}
	\begin{tabular}{l@{}rrrrrrrrrrcr}
		\toprule
					& \bf Test	& \bf Test 	& \bf Provide	& \bf Request	& \bf Signal	& \bf Signal	& \bf Provide	& \bf Provide	& \bf Provide	& \bf Other && \bf Macro	\\
		\bf  Approach 	& \bf U.\ ($e_1$) & \bf P.K.\ ($e_2$) & \bf  Ex.\ ($e_3$) & \bf Ex.\ ($e_4$) & \bf U.\ ($e_5$) & \bf N.U.\ ($e_6$) & \bf Fe.\ ($e_7$) & \bf As.\ ($e_8$) & \bf E.I.\ ($e_9$) & ($e_{10}$)  && \bf F$_1$-Score \\
		\midrule		
		BERT-basic 	   & \bf 0.27 & \bf 0.64 & \bf 0.84 & 0.60 & 0.29 & \bf 0.34 & 0.51 & 0.00 & \bf 0.11 & 0.50 && 0.41  \\
		BERT-sequence  & \bf 0.27 & \bf 0.64 & \bf 0.84 & \bf0.64 & \bf 0.33 & 0.21 & \bf **0.60 & \bf 0.15 & 0.08 & \bf 0.56 && \bf 0.43\\
		BERT-multitask & 0.21 & 0.54 & 0.80 & 0.40 & 0.16 & 0.32 & 0.53 & 0.00 & 0.08 & 0.35 && 0.34	\\
		\addlinespace
		Majority baseline   & 0.00 & 0.00 & 0.61 & 0.00 & 0.00 & 0.00 & 0.00 & 0.00 & 0.00 & 0.00 && 0.06 \\
		\bottomrule
	\end{tabular}
	\caption{Explanation move prediction results: The F$_1$-scores of the evaluated BERT models for each considered explanation move, $e_1$--$e_{10}$, as well as the macro-averaged F$_1$-score. The best value in each column is marked bold.}
	\label{table-results-explanation-move}
\end{table*}

\paragraph{BERT-multitask} 

Finally, the interaction of topic $t_i$, act $d_i$, and move $e_i$ in a turn may be relevant. For example, an \emph{informing statement} likely \emph{provides an explanation} (see Table~\ref{table-pairs}). Our third model thus learns to classify all three dimensions jointly in a multitask fashion, based on multitask-NLP.%
\footnote{Multitask NLP, https://multi-task-nlp.readthedocs.io}
We trained one multitask model each with one of the three dimensions as main task and the others as auxiliary tasks, oversampling with respect to the main task. To this end, we employ a shared BERT encoder and three classification heads, one for each task. The final loss is the weighted average of the three classification losses, with weight 0.5 for the main task and 0.25 for both others. We trained the models for 10 epochs allowing them to converge.

\subsection{Results}

Tables~\ref{table-results-topic}--\ref{table-results-explanation-move} show the individual and the macro F$_1$-scores for all three dimensions. 

\emph{BERT-sequence} performs best across all three labeling tasks, highlighting the impact of modeling the sequential interaction in dialogues. It achieves a macro F$_1$-score of 0.52 for topics, 0.47 for dialogue acts, and 0.43 for explanation moves. However, likely due to data sparsity, some labels remain hard to predict, such as \emph{Subtopic} (t$_2$), \emph{disagreement statements} (d$_8$), and \emph{provide assessment} (e$_8$).

\emph{BERT-basic} beats BERT-sequence on a few labels, such as \emph{signal non-understanding} (e$_8$), but cannot compete overall. \emph{BERT-multitask} performs worst among the three models. We attribute this to the data imbalance: While oversampling helps with respect to the main task, it does not benefit the label distribution of the auxiliary tasks. Also, optimizing the loss weights of the three tasks may~further aid multitask learning, but such an engineering of prediction models is not the focus of this work.


\section{Conclusion}
\label{sec:conclusion}

How humans explain in dialogical settings is still understudied. This paper has presented a first corpus for computational research on controlled explaining dialogues, manually annotated for topics, dialogue acts, and explanation moves. Our analysis has revealed intuitive differences in the language of explainers and explainees and their dependence on the explainee's proficiency. Moreover, baseline experiments suggest that a prediction of the annotated dimensions is feasible and benefits from modeling interactions. 
With these results, we lay the ground towards more human-centered XAI. We expect that respective systems need to learn to how to explain depending on the explainee's reactions, and how to proactively lead an explaining dialogue to achieve understanding on the explainee's side.

A limitation of the corpus lies in the restricted corpus size caused by the availability of source data, preventing deeper statistical analyses and likely rendering a direct training of dialogue systems on the corpus hard. Also, it remains to be explored what findings generalize beyond the controlled setting of the given dialogues. Future work should thus~target both the scale and the heterogeneity of explaining data, in order to provide the pervasive communicative process of explaining the attention it deserves.\,\,



\section*{Acknowledgments}

This work has been supported by the Deutsche For\-schungs\-ge\-mein\-schaft (DFG, German Research Foundation), partially under project number TRR 318/1 2021 -- 438445824 and partially under SFB 901/3 -- 160364472. We thank Meisam Booshehri, Henrik Buschmeier, Philipp Cimiano, Josephine Fisher, Angela Grimminger, and Erick Ronoh for their input and feedback to the annotation scheme. We also thank Akshit Bhatia for his help with the corpus preparation as well as the anonymous freelancers on Upwork for their annotations.


\section{Ethical Statement}
\label{sec:ethics}

We do not see any immediate ethical concerns with respect to the research in this paper. The data included in the corpus is freely available. All participants involved in the given dialogues gave their consent to be recorded and received expense allowances, as far as perceivable from the Wired web resources. As discussed in the paper, the three freelancers in our annotation study were paid about \$13 per hour, which exceeds the minimum wage in most US states and is also conform to the standards in the regions of our host institution. In our view, the provided prediction models target dimensions of dialogue turns that are not prone to be misused for ethically doubtful applications.

\bibliography{coling22-explainability-corpus-lit}

\begin{thebibliography}{33}
\expandafter\ifx\csname natexlab\endcsname\relax\def\natexlab#1{#1}\fi

\bibitem[{Al~Khatib et~al.(2016)Al~Khatib, Wachsmuth, Kiesel, Hagen, and
  Stein}]{alkhatib:2016b}
Khalid Al~Khatib, Henning Wachsmuth, Johannes Kiesel, Matthias Hagen, and Benno
  Stein. 2016.
\newblock \href {http://www.aclweb.org/anthology/C16-1324} {A news editorial
  corpus for mining argumentation strategies}.
\newblock In \emph{Proceedings of COLING 2016, the 26th International
  Conference on Computational Linguistics: Technical Papers}, pages 3433--3443.
  The COLING 2016 Organizing Committee.

\bibitem[{Al~Khatib et~al.(2018)Al~Khatib, Wachsmuth, Lang, Herpel, Hagen, and
  Stein}]{alkhatib:2018a}
Khalid Al~Khatib, Henning Wachsmuth, Kevin Lang, Jakob Herpel, Matthias Hagen,
  and Benno Stein. 2018.
\newblock \href {http://aclweb.org/anthology/P18-1237} {Modeling deliberative
  argumentation strategies on wikipedia}.
\newblock In \emph{Proceedings of the 56th Annual Meeting of the Association
  for Computational Linguistics (Volume 1: Long Papers)}, pages 2545--2555.
  Association for Computational Linguistics.

\bibitem[{{Barredo Arrieta} et~al.(2020){Barredo Arrieta},
  D\'{i}az-Rodr\'{i}guez, {Del Ser}, Bennetot, Tabik, Barbado, Garcia,
  Gil-Lopez, Molina, Benjamins, Chatila, and Herrera}]{barriedoarrieta:2020}
Alejandro {Barredo Arrieta}, Natalia D\'{i}az-Rodr\'{i}guez, Javier {Del Ser},
  Adrien Bennetot, Siham Tabik, Alberto Barbado, Salvador Garcia, Sergio
  Gil-Lopez, Daniel Molina, Richard Benjamins, Raja Chatila, and Francisco
  Herrera. 2020.
\newblock \href {https://doi.org/https://doi.org/10.1016/j.inffus.2019.12.012}
  {Explainable artificial intelligence {(XAI)}: {C}oncepts, taxonomies,
  opportunities and challenges toward responsible {AI}}.
\newblock \emph{Information Fusion}, 58:82--115.

\bibitem[{Bourse and Saint-Dizier(2012)}]{bourse:2012}
Sarah Bourse and Patrick Saint-Dizier. 2012.
\newblock \href
  {http://www.lrec-conf.org/proceedings/lrec2012/pdf/137_Paper.pdf} {A
  repository of rules and lexical resources for discourse structure analysis:
  the case of explanation structures}.
\newblock In \emph{Proceedings of the Eighth International Conference on
  Language Resources and Evaluation ({LREC}-2012)}, pages 2778--2785, Istanbul,
  Turkey. European Languages Resources Association (ELRA).

\bibitem[{Bunt et~al.(2010)Bunt, Alexandersson, Carletta, Choe, Fang, Hasida,
  Lee, Petukhova, Popescu-Belis, Romary, Soria, and Traum}]{bunt:2010}
Harry Bunt, Jan Alexandersson, Jean Carletta, Jae-Woong Choe, Alex~Chengyu
  Fang, Koiti Hasida, Kiyong Lee, Volha Petukhova, Andrei Popescu-Belis,
  Laurent Romary, Claudia Soria, and David Traum. 2010.
\newblock \href
  {http://www.lrec-conf.org/proceedings/lrec2010/pdf/560_Paper.pdf} {Towards an
  {ISO} standard for dialogue act annotation}.
\newblock In \emph{Proceedings of the Seventh International Conference on
  Language Resources and Evaluation ({LREC}'10)}, Valletta, Malta. European
  Language Resources Association (ELRA).

\bibitem[{Confalonieri et~al.(2019)Confalonieri, Besold, Weyde, Creel,
  Lombrozo, Mueller, and Shafto}]{confalonieri:2019}
Roberto Confalonieri, Tarek~R. Besold, Tillman Weyde, Kathleen Creel, Tania
  Lombrozo, Shane~T. Mueller, and Patrick Shafto. 2019.
\newblock \href {https://mindmodeling.org/cogsci2019/papers/0013/index.html}
  {What makes a good explanation? {C}ognitive dimensions of explaining
  intelligent machines}.
\newblock In \emph{Proceedings of the 41th Annual Meeting of the Cognitive
  Science Society, CogSci 2019: Creativity + Cognition + Computation, Montreal,
  Canada, July 24-27, 2019}, pages 25--26.

\bibitem[{Devlin et~al.(2019)Devlin, Chang, Lee, and Toutanova}]{devlin:2019}
Jacob Devlin, Ming-Wei Chang, Kenton Lee, and Kristina Toutanova. 2019.
\newblock \href {https://doi.org/10.18653/v1/N19-1423} {{BERT}: {P}re-training
  of deep bidirectional transformers for language understanding}.
\newblock In \emph{Proceedings of the 2019 {Conference} of the {North}
  {American} {Chapter} of the {Association} for {Computational} {Linguistics}:
  {Human} {Language} {Technologies}, {Volume} 1 ({Long} and {Short} {Papers})},
  pages 4171--4186, Minneapolis, Minnesota. Association for Computational
  Linguistics.

\bibitem[{Dulceanu et~al.(2018)Dulceanu, Le~Dinh, Chang, Bui, Kim, Vu, and
  Kim}]{dulceanu:2018}
Andrei Dulceanu, Thang Le~Dinh, Walter Chang, Trung Bui, Doo~Soon Kim,
  Manh~Chien Vu, and Seokhwan Kim. 2018.
\newblock \href {https://www.aclweb.org/anthology/L18-1438}
  {{P}hotoshop{Q}ui{A}: {A} corpus of non-factoid questions and answers for
  why-question answering}.
\newblock In \emph{Proceedings of the Eleventh International Conference on
  Language Resources and Evaluation ({LREC} 2018)}, Miyazaki, Japan. European
  Language Resources Association (ELRA).

\bibitem[{Dzikovska et~al.(2012)Dzikovska, Nielsen, and Brew}]{dzikovska:2012}
Myroslava~O. Dzikovska, Rodney~D. Nielsen, and Chris Brew. 2012.
\newblock \href {https://www.aclweb.org/anthology/N12-1021} {Towards effective
  tutorial feedback for explanation questions: A dataset and baselines}.
\newblock In \emph{Proceedings of the 2012 Conference of the North {A}merican
  Chapter of the Association for Computational Linguistics: Human Language
  Technologies}, pages 200--210, Montr{\'e}al, Canada. Association for
  Computational Linguistics.

\bibitem[{Fan et~al.(2019)Fan, Jernite, Perez, Grangier, Weston, and
  Auli}]{fan:2019}
Angela Fan, Yacine Jernite, Ethan Perez, David Grangier, Jason Weston, and
  Michael Auli. 2019.
\newblock \href {https://doi.org/10.18653/v1/P19-1346} {{ELI}5: {L}ong form
  question answering}.
\newblock In \emph{Proceedings of the 57th Annual Meeting of the Association
  for Computational Linguistics}, pages 3558--3567, Florence, Italy.
  Association for Computational Linguistics.

\bibitem[{Finke et~al.(2022)Finke, Horwath, Matzner, and Schulz}]{finke:2022}
Josefine Finke, Ilona Horwath, Tobias Matzner, and Christian Schulz. 2022.
\newblock (de)coding social practice in the field of xai: {T}owards a
  co-constructive framework of explanations and understanding between lay users
  and algorithmic systems.
\newblock In \emph{Artificial Intelligence in HCI}, pages 149--160, Cham.
  Springer International Publishing.

\bibitem[{Fontan and Saint-Dizier(2008)}]{fontan:2008}
Lionel Fontan and Patrick Saint-Dizier. 2008.
\newblock \href {https://aclanthology.org/W08-2210} {Analyzing the explanation
  structure of procedural texts: Dealing with advice and warnings}.
\newblock In \emph{Semantics in Text Processing. {STEP} 2008 Conference
  Proceedings}, pages 115--127. College Publications.

\bibitem[{Fried et~al.(2018)Fried, Andreas, and Klein}]{fried:2018}
Daniel Fried, Jacob Andreas, and Dan Klein. 2018.
\newblock \href {https://doi.org/10.18653/v1/N18-1177} {Unified pragmatic
  models for generating and following instructions}.
\newblock In \emph{Proceedings of the 2018 Conference of the North {A}merican
  Chapter of the Association for Computational Linguistics: Human Language
  Technologies, Volume 1 (Long Papers)}, pages 1951--1963, New Orleans,
  Louisiana. Association for Computational Linguistics.

\bibitem[{Garfinkel(2009)}]{garfinkel:2009}
Alan Garfinkel. 2009.
\newblock \emph{Forms of {Explanation}: {Rethinking} the {Questions} in
  {Social} {Theory}}, revised edition.
\newblock Yale University Press, New Haven \& London, New Haven; London.

\bibitem[{Gilpin et~al.(2018)Gilpin, Bau, Yuan, Bajwa, Specter, and
  Kagal}]{gilpin:2018}
Leilani~H. Gilpin, David Bau, Ben~Z. Yuan, Ayesha Bajwa, Michael Specter, and
  Lalana Kagal. 2018.
\newblock \href {http://arxiv.org/abs/1806.00069} {Explaining explanations:
  {A}n overview of interpretability of machine learning}.
\newblock ArXiv: 1806.00069.

\bibitem[{Goodman and Flaxman(2017)}]{goodman:2017}
Bryce Goodman and Seth Flaxman. 2017.
\newblock \href {https://doi.org/10.1609/aimag.v38i3.2741} {European union
  regulations on algorithmic decision-making and a ``right to explanation''}.
\newblock \emph{AI Magazine}, 38(3):50--57.

\bibitem[{Habernal et~al.(2018)Habernal, Wachsmuth, Gurevych, and
  Stein}]{habernal:2018a}
Ivan Habernal, Henning Wachsmuth, Iryna Gurevych, and Benno Stein. 2018.
\newblock \href {http://aclweb.org/anthology/N18-1175} {The argument reasoning
  comprehension task: Identification and reconstruction of implicit warrants}.
\newblock In \emph{Proceedings of the 2018 Conference of the North American
  Chapter of the Association for Computational Linguistics: Human Language
  Technologies, Volume 1 (Long Papers)}, pages 1930--1940. Association for
  Computational Linguistics.

\bibitem[{Hovy et~al.(2013)Hovy, Berg-Kirkpatrick, Vaswani, and
  Hovy}]{hovy:2013}
Dirk Hovy, Taylor Berg-Kirkpatrick, Ashish Vaswani, and Eduard Hovy. 2013.
\newblock \href {https://aclanthology.org/N13-1132} {Learning whom to trust
  with {MACE}}.
\newblock In \emph{Proceedings of the 2013 Conference of the North {A}merican
  Chapter of the Association for Computational Linguistics: Human Language
  Technologies}, Atlanta, Georgia. Association for Computational Linguistics.

\bibitem[{Jansen et~al.(2016)Jansen, Balasubramanian, Surdeanu, and
  Clark}]{jansen:2016}
Peter Jansen, Niranjan Balasubramanian, Mihai Surdeanu, and Peter Clark. 2016.
\newblock \href {https://aclanthology.org/C16-1278} {What{'}s in an
  explanation? characterizing knowledge and inference requirements for
  elementary science exams}.
\newblock In \emph{Proceedings of {COLING} 2016, the 26th International
  Conference on Computational Linguistics: Technical Papers}, pages 2956--2965,
  Osaka, Japan. The COLING 2016 Organizing Committee.

\bibitem[{Jordan et~al.(2006)Jordan, Makatchev, and Pappuswamy}]{jordan:2006}
Pamela~W. Jordan, Maxim Makatchev, and Umarani Pappuswamy. 2006.
\newblock \href {https://aclanthology.org/W06-3503} {Understanding complex
  natural language explanations in tutorial applications}.
\newblock In \emph{Proceedings of the Third Workshop on Scalable Natural
  Language Understanding}, pages 17--24, New York City, New York. Association
  for Computational Linguistics.

\bibitem[{Li et~al.(2021)Li, Zhang, and Chen}]{li:2021}
Lei Li, Yongfeng Zhang, and Li~Chen. 2021.
\newblock \href {https://doi.org/10.18653/v1/2021.acl-long.383} {Personalized
  transformer for explainable recommendation}.
\newblock In \emph{Proceedings of the 59th Annual Meeting of the Association
  for Computational Linguistics and the 11th International Joint Conference on
  Natural Language Processing (Volume 1: Long Papers)}, pages 4947--4957,
  Online. Association for Computational Linguistics.

\bibitem[{Mann and Thompson(1988)}]{mann:1988}
William~C Mann and Sandra~A Thompson. 1988.
\newblock Rhetorical structure theory: {T}oward a functional theory of text
  organization.
\newblock \emph{Text-interdisciplinary Journal for the Study of Discourse},
  8(3):243--281.

\bibitem[{Miller(2019)}]{miller:2019}
Tim Miller. 2019.
\newblock \href {https://doi.org/10.1016/j.artint.2018.07.007} {Explanation in
  artificial intelligence: {Insights} from the social sciences}.
\newblock \emph{Artificial Intelligence}, 267:1--38.

\bibitem[{Nakov et~al.(2017)Nakov, Hoogeveen, M{\`a}rquez, Moschitti, Mubarak,
  Baldwin, and Verspoor}]{nakov:2017}
Preslav Nakov, Doris Hoogeveen, Llu{\'\i}s M{\`a}rquez, Alessandro Moschitti,
  Hamdy Mubarak, Timothy Baldwin, and Karin Verspoor. 2017.
\newblock \href {https://doi.org/10.18653/v1/S17-2003} {{S}em{E}val-2017 task
  3: Community question answering}.
\newblock In \emph{Proceedings of the 11th International Workshop on Semantic
  Evaluation ({S}em{E}val-2017)}, pages 27--48, Vancouver, Canada. Association
  for Computational Linguistics.

\bibitem[{Rohlfing et~al.(2021)Rohlfing, Cimiano, Scharlau, Matzner, Buhl,
  Buschmeier, Esposito, Grimminger, Hammer, H\"{a}b-Umbach, Horwath,
  H\"{u}llermeier, Kern, Kopp, Thommes, Ngonga~Ngomo, Schulte, Wachsmuth,
  Wagner, and Wrede}]{rohlfing:2021}
Katharina~J. Rohlfing, Philipp Cimiano, Ingrid Scharlau, Tobias Matzner,
  Heike~M. Buhl, Hendrik Buschmeier, Elena Esposito, Angela Grimminger, Barbara
  Hammer, Reinhold H\"{a}b-Umbach, Ilona Horwath, Eyke H\"{u}llermeier,
  Friederike Kern, Stefan Kopp, Kirsten Thommes, Axel-Cyrille Ngonga~Ngomo,
  Carsten Schulte, Henning Wachsmuth, Petra Wagner, and Britta Wrede. 2021.
\newblock \href {https://doi.org/10.1109/TCDS.2020.3044366} {Explanation as a
  social practice: {T}oward a conceptual framework for the social design of ai
  systems}.
\newblock \emph{IEEE Transactions on Cognitive and Developmental Systems},
  13(3):717--728.

\bibitem[{Situ et~al.(2021)Situ, Zukerman, Paris, Maruf, and
  Haffari}]{situ:2021}
Xuelin Situ, Ingrid Zukerman, Cecile Paris, Sameen Maruf, and Gholamreza
  Haffari. 2021.
\newblock \href {https://doi.org/10.18653/v1/2021.acl-long.415} {Learning to
  explain: Generating stable explanations fast}.
\newblock In \emph{Proceedings of the 59th Annual Meeting of the Association
  for Computational Linguistics and the 11th International Joint Conference on
  Natural Language Processing (Volume 1: Long Papers)}, pages 5340--5355,
  Online. Association for Computational Linguistics.

\bibitem[{Son et~al.(2018)Son, Bayas, and Schwartz}]{son:2018}
Youngseo Son, Nipun Bayas, and H.~Andrew Schwartz. 2018.
\newblock \href {https://www.aclweb.org/anthology/D18-1372} {Causal explanation
  analysis on social media}.
\newblock In \emph{Proceedings of the 2018 Conference on Empirical Methods in
  Natural Language Processing}, pages 3350--3359, Brussels, Belgium.
  Association for Computational Linguistics.

\bibitem[{Stolcke et~al.(2000)Stolcke, Ries, Coccaro, Shriberg, Bates,
  Jurafsky, Taylor, Martin, Van Ess-Dykema, and Meteer}]{stolcke:2000}
Andreas Stolcke, Klaus Ries, Noah Coccaro, Elizabeth Shriberg, Rebecca Bates,
  Daniel Jurafsky, Paul Taylor, Rachel Martin, Carol Van Ess-Dykema, and Marie
  Meteer. 2000.
\newblock \href {https://aclanthology.org/J00-3003} {Dialogue act modeling for
  automatic tagging and recognition of conversational speech}.
\newblock \emph{Computational Linguistics}, 26(3):339--374.

\bibitem[{Swales(1990)}]{swales:1990}
John~M. Swales. 1990.
\newblock \emph{Genre Analysis: {E}nglish in Academic and Research Settings}.
\newblock Cambridge University Press.

\bibitem[{Vander~Linden(1992)}]{vander:1992}
Keith Vander~Linden. 1992.
\newblock The expression of local rhetorical relations in instructional text.
\newblock In \emph{Proceedings of the 30th Annual Meeting of the Association
  for Computational Linguistics}, pages 318--320.

\bibitem[{Wachsmuth and Stein(2017)}]{wachsmuth:2017c}
Henning Wachsmuth and Benno Stein. 2017.
\newblock \href {https://doi.org/http://doi.acm.org/10.1145/2957757} {A
  universal model for discourse-level argumentation analysis}.
\newblock \emph{Special Section of the {ACM} Transactions on Internet
  Technology: {A}rgumentation in Social Media}, 17(3):28:1--28:24.

\bibitem[{Yagcioglu et~al.(2018)Yagcioglu, Erdem, Erdem, and
  Ikizler-Cinbis}]{yagcioglu:2018}
Semih Yagcioglu, Aykut Erdem, Erkut Erdem, and Nazli Ikizler-Cinbis. 2018.
\newblock \href {https://doi.org/10.18653/v1/D18-1166} {{R}ecipe{QA}: A
  challenge dataset for multimodal comprehension of cooking recipes}.
\newblock In \emph{Proceedings of the 2018 Conference on Empirical Methods in
  Natural Language Processing}, pages 1358--1368, Brussels, Belgium.
  Association for Computational Linguistics.

\bibitem[{Zhang et~al.(2012)Zhang, Webster, Uren, Varga, and
  Ciravegna}]{zhang:2012}
Ziqi Zhang, Philip Webster, Victoria Uren, Andrea Varga, and Fabio Ciravegna.
  2012.
\newblock \href
  {http://www.lrec-conf.org/proceedings/lrec2012/pdf/244_Paper.pdf}
  {Automatically extracting procedural knowledge from instructional texts using
  natural language processing}.
\newblock In \emph{Proceedings of the Eighth International Conference on
  Language Resources and Evaluation ({LREC}'12)}, pages 520--527, Istanbul,
  Turkey. European Language Resources Association (ELRA).

\end{thebibliography}

\end{document}